\documentclass[10pt,twocolumn,letterpaper]{article}

\usepackage{iccv}
\usepackage{times}
\usepackage{epsfig}
\usepackage{graphicx}
\usepackage{amsmath}
\usepackage{amssymb}

\usepackage{multirow}
\usepackage[caption=false,font=footnotesize]{subfig}
\usepackage[bold]{hhtensor}

\newcommand{\squishlist}{
 \begin{list}{$\bullet$}
  { \setlength{\itemsep}{0pt}
     \setlength{\parsep}{3pt}
     \setlength{\topsep}{3pt}
     \setlength{\partopsep}{0pt}
     \setlength{\leftmargin}{1.5em}
     \setlength{\labelwidth}{1em}
     \setlength{\labelsep}{0.5em} } }

\newcommand{\squishend}{
  \end{list}  }

\usepackage[breaklinks=true,bookmarks=false]{hyperref}

\iccvfinalcopy 

\def\iccvPaperID{****} 

\begin{document}

\title{Weakly Supervised Fine-Grained Image Categorization}


\author{Yu Zhang$^1$ \and Xiu-shen Wei$^2$ \and Jianxin Wu$^2$  \and Jianfei Cai$^3$
\and Jiangbo Lu$^1$ \and Viet-Anh Nguyen$^1$ \and Minh N. Do$^1$ \and
{$^1$Advanced Digital Sciences Center, Singapore} \\
{\tt\small \{zhang.yu, Jiangbo.Lu, vanguyen\}@adsc.com.sg, minhdo@illinois.edu}
\and
{$^2$ National Key Laboratory for Novel Software Technology, Nanjing University, Nanjing, China} \\
{\tt\small weixs@lamda.nju.edu.cn, wujx2001@nju.edu.cn}
\and
{$^3$ School of Computer Engineering, Nanyang Technological University, Singapore}\\
{\tt\small asjfcai@ntu.edu.sg}
}

\maketitle

\begin{abstract}
 In this paper, we categorize fine-grained images without using any object / part annotation neither in the training nor in the testing stage, a step towards making it suitable for deployments. Fine-grained image categorization aims to classify objects with subtle distinctions. Most existing works heavily rely on object / part detectors to build the correspondence between object parts by using object or object part annotations inside training images. The need for expensive object annotations prevents the wide usage of these methods. Instead, we propose to select useful parts from multi-scale part proposals in objects, and use them to compute a global image representation for categorization. This is specially designed for the annotation-free fine-grained categorization task, because useful parts have shown to play an important role in existing annotation-dependent works but accurate part detectors can be hardly acquired. With the proposed image representation, we can further detect and visualize the key (most discriminative) parts in objects of different classes. In the experiment, the proposed annotation-free method achieves better accuracy than that of state-of-the-art annotation-free and most existing annotation-dependent methods on two challenging datasets, which shows that it is not always necessary to use accurate object / part annotations in fine-grained image categorization.
\end{abstract}

\section{Introduction}

Fine-grained image categorization has been popular during the past few years. Different from traditional image recognition such as scene or object recognition, fine-grained categorization
deals with images with subtle distinctions, which usually involves the classification of subclasses of objects belonging to the same class like birds~\cite{WahCUB_200_2011,Berg2014CVPR_Birdsnap},
dogs~\cite{KhoslaYaoJayadevaprakashFeiFei_FGVC2011}, planes~\cite{Vedaldi2014CVPR_fineplanes}, plants~\cite{Nilsback2008ICVGIP_flowers,Rejeb2013CVPR_VantageFine}, etc. Therefore, it requires methods that are more discriminative than traditional image classification.

One important common feature of existing fine-grained methods is that they  \emph{explicitly use annotations of the object or even object parts} to depict an object as precise as possible. Most of them heavily rely on object / part detectors to find the part correspondence among objects.

For example, in~\cite{Farrell2011ICCV_Birdlets,Ning2012CVPR_posepooling}, the poselet~\cite{Bourdev2010ECCV_Poselet} is used to detect object parts. Then, each object is represented with a bag of poselets, and suitable matchings among poselets (parts) could be found between two objects. Instead of using poselets, \cite{Ning2013CVPR_DPD} used the deformable part models (DPM)~\cite{Felzenszwalb2010PAMI_DPM} for object part detection. DPM is learned from the annotated object parts in training objects, which is then applied on testing objects to detect parts. Some works~\cite{Gavves2013ICCV_finealign,Christoph2014CVPR_Part} transfer the part annotations from objects in training images to those sharing similar shapes in testing images instead of applying object / part detectors. Instead of seeking precise part localization, \cite{Gavves2013ICCV_finealign} also provided an unsupervised object alignment technique, which roughly aligns objects and divides them into corresponding parts along certain directions. It achieves better results than the label transfer method. Recently, \cite{ning2014ECCV_cnnFine-grained} proposed to use object and part detectors with powerful CNN feature representations~\cite{Donahue2014ICML_decaf}, which achieves state-of-the-art results on the Caltech-UCSD Birds (CUB) 200-2011~\cite{WahCUB_200_2011} dataset. The geometric relation between an object and its parts are considered in~\cite{ning2014ECCV_cnnFine-grained}. \cite{Ning2014CVPR_PANDA} also shows that part-based models with CNN features is able to capture subtle distinctions among objects. Some other works~\cite{Jia2013CVPR_FineCrowdsourcing,Catherine2014CVPR_Interactivefine} recognize fine-grained images with human in the loop.

In order to achieve accurate part detection, most existing works require the annotated bounding boxes for objects, in both training and testing stages. As pointed out in~\cite{ning2014ECCV_cnnFine-grained}, such a requirement is not so realistic for practical usage. Thus, a few works~\cite{ning2014ECCV_cnnFine-grained,Donahue2014ICML_decaf} have looked into a more realistic setup, i.e., only utilize the bounding box in the training stage but not in the testing stage. However, even with such a setup, it is still hard for wide deployment of these methods since accurate object annotations needed in the training stage are usually expensive to acquire, especially for large-scale image classification problems. It is an interesting research problem to free us from the dependency to detailed manual annotations in fine-grained image categorization tasks. \cite{Tianjun2014arxiv_fine2cnn} has shown promising results without using manual annotations. They try to detect accurate objects and parts with complex deep learning models for fine-grained recognition.

\begin{figure}
 \centering
 \includegraphics[width=0.475\textwidth]{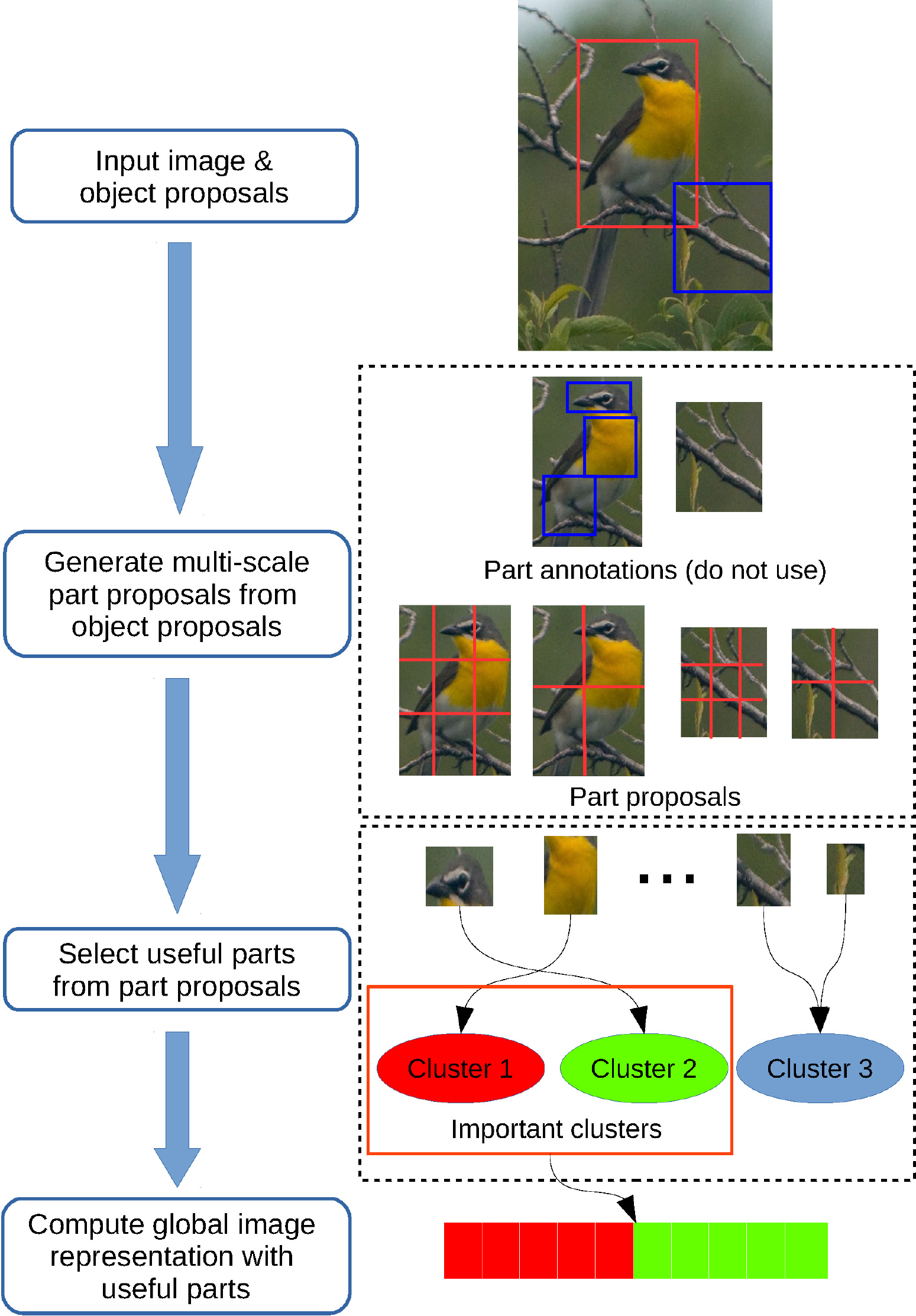}
 \caption{System overview. This figure is best viewed in color.} \label{fig_system}
\end{figure}

In this paper, we aim at \emph{categorizing fine-grained images with only category labels and without any bounding box annotation in both training and testing stages, while not degrading the categorization accuracy}. This is a big step towards making fine-grained image categorization suitable for wide deployments. In existing annotation-dependent works, representative parts like head and body in birds~\cite{ning2014ECCV_cnnFine-grained} have shown to play the key role in capturing the subtle difference of fine-grained images. They are different from general image recognition, which usually uses a holistic image representation. In this paper, we are going to select the most important parts from multiple part proposals in each image in the annotation-free scenario. The part proposals are the sub-regions of object proposals in each image, which are shown in Fig.~\ref{fig_system}. The part selection process is important in an annotation-free fine-grained image categorization system for at least two reasons. First, many (if not most) part proposals are noise and not useful for categorization. Second, accurate part detectors can be hardly acquired without access to detailed object and part annotations, including groundtruth exact object and part locations. Existing accurate part detectors (e.g.,~\cite{ning2014ECCV_cnnFine-grained} ) are annotation-dependent, different from our annotation-free setup.

We propose to select \textit{many} useful parts from multi-scale part proposals of objects in each image and compute a global image representation for it, which is then used to learn a linear classifier for image categorization. In this image representation, we believe that to select many useful parts is better than one exact part, because it is very difficult to determine the exact object / part location in the image in our annotation-free scenario. Multiple useful parts can compensate each other to provide more useful information in characterizing the object. The proposed representation achieves better accuracy than the annotation-free work~\cite{Tianjun2014arxiv_fine2cnn} and even most existing annotation-dependent methods on two challenging benchmark datasets, the Caltech-UCSD Birds 200-2011~\cite{WahCUB_200_2011} and the StanfordDogs~\cite{KhoslaYaoJayadevaprakashFeiFei_FGVC2011} datasets. Its success suggests that it is not always necessary to learn expensive object / part detectors in fine-grained image categorization.

Fig.~\ref{fig_system} gives an overview on how we generate a global representation for each image through selected parts. The framework in Fig.~\ref{fig_system} consists of three major steps: \emph{part proposal generation}, \emph{useful part selection}, and \emph{multi-scale image representation}.

\squishlist
 \item In the first step, we extract object proposal which are image patches that may contain an object. Then, multi-scale part proposals are extracted from object proposals in each image. An efficient multi-max pooling (MMP) strategy is proposed to generate features for multi-scale part proposals by leveraging the internal structure of CNN on object proposals. Within the large number of part proposals, most of them are from background clutters, which are harmful to categorization.

 \item Thus, in the second step, we select useful part proposals from each image by exploring useful information in part clusters (all part proposals are clustered). For each part cluster, we compute an importance score for it, indicating how important is this cluster for our fine-grained task. Those part proposals assigned to useful clusters (i.e., those with largest importance scores) are selected as useful parts.

 \item Finally, the selected part proposals in each image are encoded into a global image representation. In order to highlight the subtle distinction among fine-grained objects, we encode the selected parts on different scales separately, which we name as SCale Pyramid Matching (ScPM). It provides a better discriminance than encoding all parts altogether in one image.
\squishend

\begin{figure}
 \centering
 \includegraphics[width=0.475\textwidth]{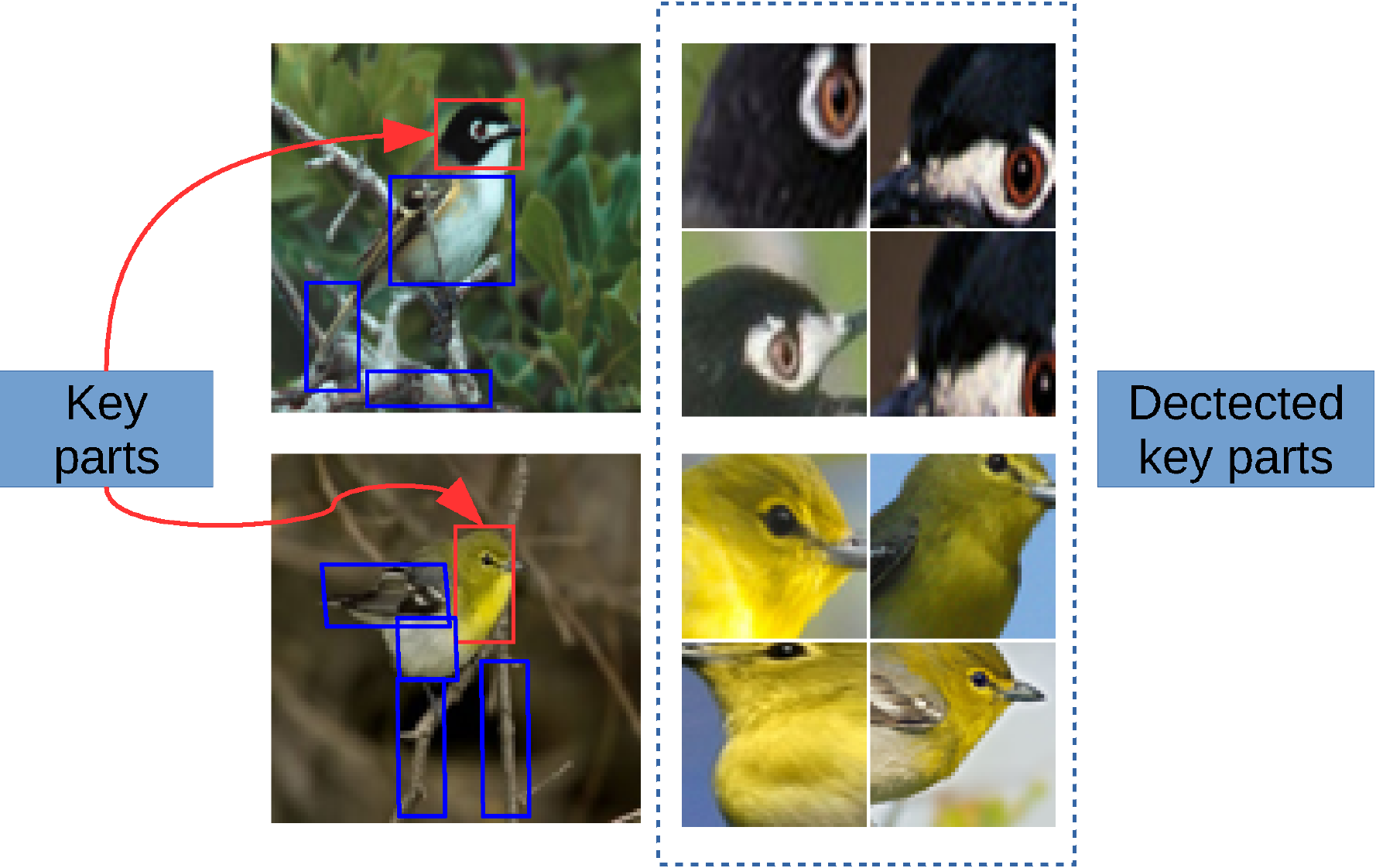}
 \caption{Black-capped Vireo and Yellow-throated Vireo. They have the most distinctive parts in multiple part proposals: yellow throat and black cap, respectively, which are specified in red boxes. On the right, we show the key parts detected using the proposed representation from the two species. More examples of detected discriminative parts can be found in Fig.~\ref{fig_show_key_parts}. This figure is best viewed in color.}
\label{fig_key_part_demo}
\end{figure}

With the proposed annotation-free fine-grained image representation, we can detect the key (most discriminative) parts in objects for different classes, whose results coincide well with rules used by human experts (e.g., the yellow-throated vireo and black-capped vireo differs because yellow-throated vireo has yellow throat while black-capped vireo has black head, cf. Fig.~\ref{fig_key_part_demo}).


\section{Fine-grained image representation without using object / part annotations}

The three modules in the proposed method (part proposal generation, part selection, and multi-scale image representation) are detailed in Sections~\ref{sec_mmp}--\ref{ScPM}, respectively.

\subsection{Part proposal generation} \label{sec_mmp}

\begin{figure*}
\centering
\includegraphics[width=0.85\textwidth]{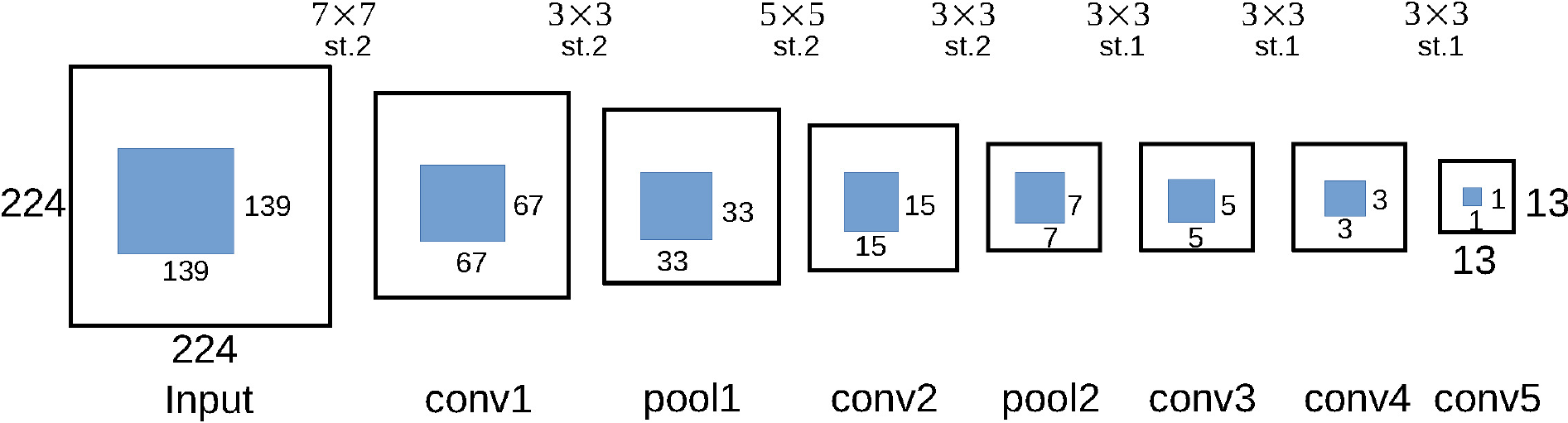}
\caption{Receptive fields computed using the CNN model (imagenet-vgg-m)~\cite{Chatfield2014BMVC_CNN}. One cell in the \textsc{conv5} layer corresponds to a $139\times 139$ receptive field in the input image. We only show the spatial size of the image and filters. $a\times a$ is the filter (pooling) size, `st' means the step size.}
\label{fig_recepfield}
\end{figure*}

Regional information has been shown to improve image classification with hand-crafted methods like spatial pyramid matching~\cite{Lazebnik2006CVPR_SPM} and receptive fields~\cite{Yangqing2012CVPR_ReceptiveField}. When a CNN model is applied on an image, features of local regions can be acquired automatically from its internal structure. Assume the output from a layer in CNN is $N\times N \times d$ dimension, which is the output of $d$ filters for $N\times N$ spatial cells. Each spatial cell is computed from a receptive field in the input image. The receptive fields of all the spatial cells in the input image can highly overlap with each other. The size of one receptive field can be computed layer by layer in CNN. In a convolution (pooling) layer, if the filter (pooling) size is $a\times a$ and the step size is $s$, then $T\times T$ cells in the output of this layer corresponds to $[s(T-1)+a]\times [s(T-1)+a]$ cells in the input of this layer. For example, one cell in the \textsc{conv5} (the 5th convolutional) layer of CNN model (imagenet-vgg-m)~\cite{Chatfield2014BMVC_CNN} corresponds to a $139\times 139$ receptive field (it is assumed to reside in the image completely) in the $224\times 224$ input image (cf. Fig.~\ref{fig_recepfield}).

The spatial cells in one CNN layer correspond to receptive fields with a fixed size, which are not comprehensive enough to characterize objects of different scales in images. Some efforts have been made to solve this problem. \cite{Wanli2014arxiv_deepidnet} applied multiple convolutional filters of different sizes in the \textsc{conv5} layer of CNN to generate multi-scale part mappings for object detection. \cite{Donggeun2014arxiv_fvcnn} applied CNN model on images resized to different scales and pool the final outputs using the Fisher vector method. These methods, however, are more time consuming than the original CNN computations.

\begin{figure}
\centering
\includegraphics[width=0.45\textwidth]{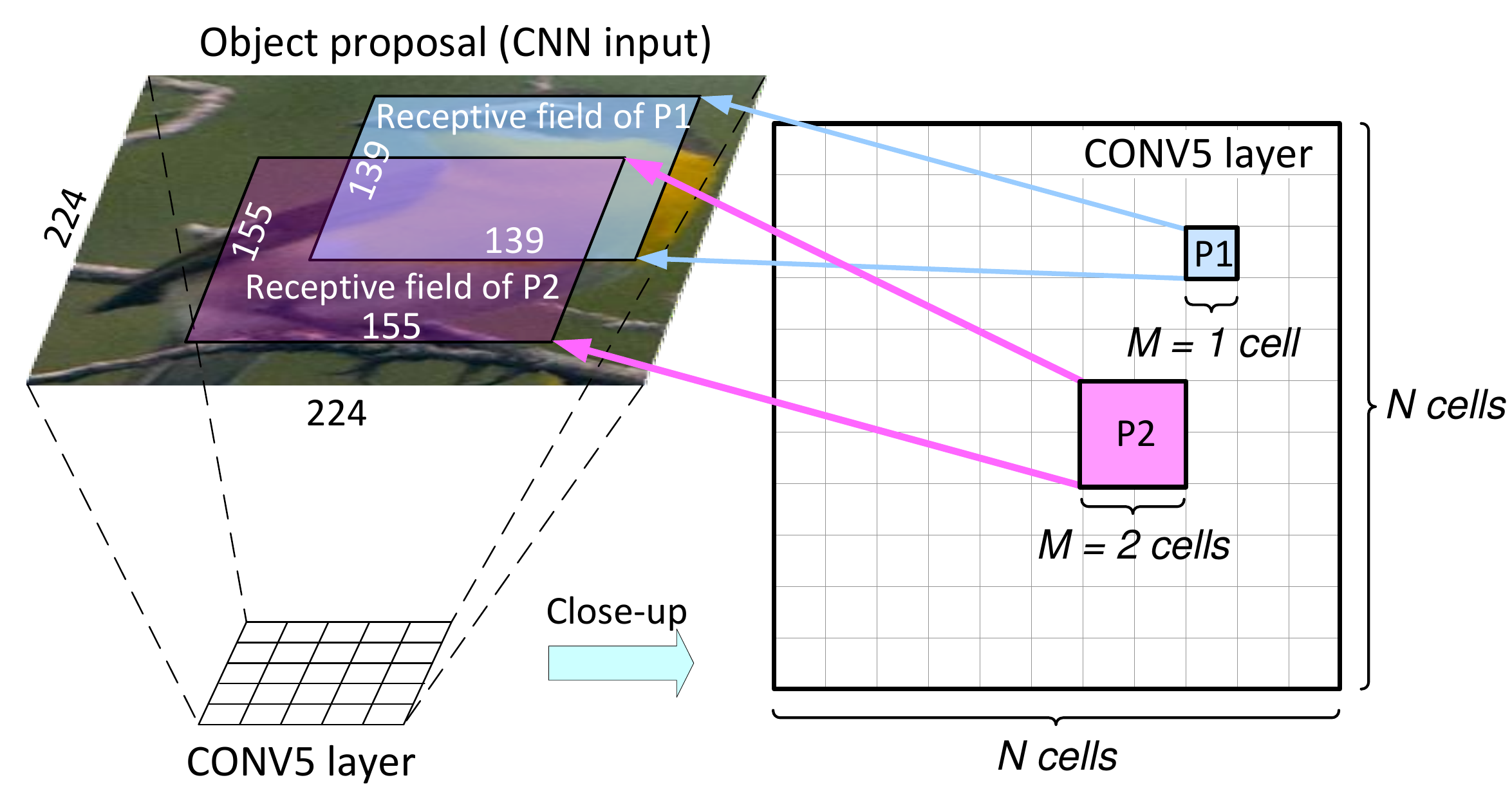}
\caption{Generating multi-scale part proposals. The input is an object proposal. By applying CNN on it, spatial cells of different sizes on the \textsc{conv5} layer in CNN correspond to parts of different scales. This figure is best viewed in color.}
\label{fig_multiscaleparts}
\end{figure}

We generate features of multi-scale receptive fields for an image by leveraging the internal outputs of CNN with little additional computational cost (cf. Fig.~\ref{fig_multiscaleparts}). Considering the outputs of one layer in a CNN, we can pool the activation vectors of adjacent cells of different sizes, which corresponds to receptive fields with different sizes in the input image. Max-pooling is used here.

Given the $N\times N$ cells in one layer in CNN, we use max-pooling to combine information from all $M\times M$ adjacent cells, where $M$ ranges from 1 (single cell) to $N$ (all the cells). When $M$ is assigned to different values, the corresponding cells can cover receptive fields of different sizes in the input image, thus providing more comprehensive information. We name the proposed part proposal generation strategy as multi-max pooling (MMP) and apply it to the \textsc{conv5} layer, because the \textsc{conv5} layer can capture more meaningful object / part information than those shallow layers in CNN~\cite{Zeiler2014ECCV_visualizeCNN}. When a CNN model is applied on an object bounding box in an image, the acquired receptive fields from MMP can be seen as the part candidates for an object, which provides a comprehensive understanding for fine-grained objects.

Part proposals are important for fine-grained image categorization, which can provide fine-grained information of objects. Object proposals generated by objectness methods like selective search~\cite{Uijlings2013IJCV_SS} are not fine-grained enough to characterize the internal structure of fine-grained objects. \cite{ning2014ECCV_cnnFine-grained} only used such object proposals to detect parts with the help of object / part annotations, which leads to worse performance than our annotation-free method using part proposals (cf. Table~\ref{table_acc_CUB200_2011}).

MMP is an efficient way to generate multi-scale part proposals to characterize fine-grained objects. It can be easily applied on millions or even billions object proposals in a dataset. Unlike~\cite{Tianjun2014arxiv_fine2cnn}, where the outputs of \textsc{conv4} in CNN are used as parts, MMP provides dense coverage on different scales from part level to object level for each object proposal. The large number of part proposals provide us more opportunity to mine subtle useful information of objects.

\subsection{Part selection}

We propose to select useful parts from global image representations. To get the image representation, we first generate object proposals from each image. Since no object / part annotations are provided, we could only use unsupervised object detection methods. Considering the efficiency, selective search~\cite{Uijlings2013IJCV_SS} is used in our framework, which has also been used in~\cite{Girshick2014CVPR_RCNN,ning2014ECCV_cnnFine-grained} to generate initial object / part candidates for object detectors. After generating multiple object proposals, we apply CNN model on each detected bounding box / object proposal, and use the proposed MMP to get part proposals.

Among the object proposals, most of them are from background clutters, which are harmful for image recognition. For example, in the CUB200-2011~\cite{WahCUB_200_2011} dataset, when we use the intersection over union criteria, only $10.4\%$ object proposals cover the foreground object. The part proposals from those unsuccessful object proposals will contribute little to the classification, or even be noisy and harmful. Thus, we need to select those useful part proposals (covering the foreground object) but without using groundtruth annotations for our image representation.

We select useful parts through mining the useful information in part clusters. We first cluster all part proposals into several groups. Then, we compute the importance of each cluster for image classification. Those part proposals assigned to the useful clusters (clusters with largest importance values) are selected as the useful parts.

We compute the cluster importance with the aid of Fisher vector (FV)~\cite{Sanchez2013IJCV_FV}.\footnote{VLAD can be used in our framework, which is used in~\cite{Yunchao2014ECCV_cnnvlad} to encode CNN of multiple spatial regions for general image classification.  We choose FV because it has a better discriminance than VLAD~\cite{Jegou2012PAMI_VLAD}.} We first encode all the part proposals in each image into a FV. Then, for each dimension in FVs of all training images, we compute its importance using its mutual information (MI) with the class labels~\cite{me2014cvpr_fs}. Finally, the cluster importance is the summation of the MI values of all FV dimensions in it. We only keep those FV dimensions from the most important clusters for image categorization.

To the best of our knowledge, explicit part proposal and selection is for the first time proposed for fine-grained image categorization, in an annotation-free setup. As will be shown in Sec.~\ref{sec_experiments}, this novel strategy greatly improves categorization accuracy, even when object or part annotations are not used at all.
Part selection can automatically explore those parts which are important for categorization by using image labels. It is more efficient and practical than trying to learn explicit part detectors without groundtruth object / part annotations.
\cite{Tianjun2014arxiv_fine2cnn} also worked on fine-grained categorization without object / part annotations, which costs much more computation than ours. \cite{Tianjun2014arxiv_fine2cnn} used two CNN models to detect interesting objects and further learn accurate part detectors from them. In contrast, we only need to select important parts from all part proposals, which are generated by applying one CNN model on object proposals.
More importantly, our method shows that without explicitly detecting the fine-grained objects / parts, the proposed image representation can acquire a better discriminance than~\cite{Tianjun2014arxiv_fine2cnn} (cf. Table~\ref{table_acc_CUB200_2011}).

\subsection{Multi-scale image representation} \label{ScPM}

We select important part clusters for parts on different scales separately. Aggregating part proposals from different scales altogether into a single image representation cannot highlight the subtle distinction in fine-grained images. Thus, we propose to encode part proposals in an image on different scales separately and we name it as SCale Pyramid Matching (ScPM). For part proposals on different scales, we compute separate FVs. In practice, the scale number can be very large ($N=13$ in the CNN model in our paper), which may lead to severe memory problem. Since the part proposals on neighboring scales are similar in size, we can divide all the scales into $m ~(m\leq N)$ groups $\{g(j), j=1,\dots,m, g(j)\subseteq \{1,\dots,N\}\}$. For an image $I$, the part proposals belonging to the scale group $g(j)$ are used to compute one FV $\phi_j(I)$ as the following:
\begin{align}
 \phi_j(I) =& [f_{\vec{\mu}_1^j}(I),f_{\vec{\sigma}_1^j}(I),\dots, f_{\vec{\mu}_i^j}(I),f_{\vec{\sigma}_i^j}(I),\dots],  \\
 f_{\vec{\mu}_i^j}(I) =& \frac{1}{\sqrt{w_i^j}}\sum_{c(t)\in g(j)}\gamma_{t}^j(i)\left(\frac{\vec{x}_t-\vec{\mu}_i^j}{\vec{\sigma}_i^j}\right),   \\
 f_{\vec{\sigma}_i^j}(I) =& \frac{1}{\sqrt{2w_i^j}}\sum_{c(t)\in g(j)}\gamma_{t}^j(i)\left[\frac{(\vec{x}_t-\vec{\mu}_i^j)^2}{(\vec{\sigma}_i^j)^2}-1\right],
\end{align}
where $\{w_i^j, \vec{\mu}_i^j, \vec{\sigma}_i^j\}$ are
respectively the mixture weights, mean vectors, and standard
deviation vectors of the $i$-th selected diagonal Gaussian in the $j$-th scale group $g(j), j=1,\dots,m$. $\{\vec{x}_t\}$ are the selected part proposals in an image, $c(t)$ is the scale index of the $t$-th part and
$\gamma_{t}^j(i)$ is the weight of the $t$-th instance to the $i$-th
Gaussian in the $j$-th scale group. Following~\cite{Sanchez2013IJCV_FV}, two parts
corresponding to the mean and the standard deviation in each
Gaussian of FV are used. Each of the $m$ FVs is power and $\ell_2$ normalized independently, and then concatenated to represent the whole image as $\phi(I)$:\footnote{The source code will be published.}
\begin{align}
\phi(I) = [\phi_1(I),\dots, \phi_m(I)].
\label{eq_image_fv}
\end{align}

ScPM is different from the Multi-scale Pyramid Pooling (MPP) method in~\cite{Donggeun2014arxiv_fvcnn}. On one hand, MPP encodes local features from images resized on different scales into separate FVs, and aggregate all the FVs into one to represent an image. Such aggregation may not highlight the subtle difference of object parts on different scales, which is especially important  in fine-grained objects within complex backgrounds. On the other hand, ScPM automatically selects different number of important part clusters on different scales. The FV representations from all scales do not have the same length and cannot be aggregated as MPP.

\section{Understand subtle visual differences: with the help of key part detection}

We want to detect and show the key (most discriminative) parts in fine-grained images of different classes to give more insightful understanding of the critical property in objects, which may help us in feature design for fine-grained images. Note that we only have image labels in training images. In order to find the key parts in a class, we need to propagate the training image labels to parts in objects. Label propagation is also used in~\cite{Donggeun2014arxiv_fvcnn} on their feature representation to compute the object confidence map in general image recognition.

Suppose we want to interpret how yellow-throated vireo is different from black-capped vireo (illustrated in Fig.~\ref{fig_key_part_demo}), we consider a binary classification problem where yellow-throated vireo and black-capped vireo are the positive and negative classes, respectively. We will compute a score for each part to denote its importance in this binary classification. A part with the largest score means that it is essential for yellow-throated vireo, and a part with the smallest score  (i.e., the most negative score) is key to black-capped vireo.

We learn a max-margin binary classifier in each selected part cluster to compute the part score. This classifier is used to propagate the image labels to parts. In the training phase, for each selected part cluster, we aggregate the part features in one image assigned into this cluster altogether (similar to VLAD). The aggregated features of training images are $\ell_2$ normalized and then used to train a classifier with image labels. In the testing phase, given a part, its score is computed as the dot-product between the classifier in the part cluster it falls in (only consider those parts in the selected part clusters) and its feature (the CNN activation vector). In both training and testing processes, the part features are centered (i.e., minus the cluster center in each part cluster).

\section{Experiments} \label{sec_experiments}

In this section, we evaluate the proposed annotation-free method on fine-grained categorization. The selective search method~\cite{Uijlings2013IJCV_SS} with default parameters is used to generate object proposals for each image. The pre-learned CNN models~\cite{Chatfield2014BMVC_CNN} from ImageNet are used to extract feature from each object proposal as~\cite{Girshick2014CVPR_RCNN}, which has been shown to achieve state-of-the-art results. It is fine-tuned with training images and their labels. But we do not fine tune CNN on object proposals because many of them are from background clutters, which may deteriorate the CNN performance. We use the `imagenet-vgg-m' model~\cite{Chatfield2014BMVC_CNN} on object proposals, given that its efficiency and accuracy are both satisfactory.

The part proposals in each scale group are assigned into 128 clusters. Each part feature is reduced into 128 dimensions by PCA. All 13 part scales ($N=13$ in the CNN model) are divided into 8 scale groups: the first 4 scales form the first 4 groups, the subsequent 6 scales form 3 groups with 2 scales in one group, and the last 3 scales form the last scale group. This arrangement make the number of parts in each group roughly balanced. The dimension of the final image representation using FV is: $128\times2\times128\times8=262144$, from which different fractions of useful part clusters will be selected and evaluated.

We evaluate the proposed method on two benchmark fine-grained datasets:
\squishlist
 \item \textbf{CUB200-2011~\cite{WahCUB_200_2011}}: The Caltech-UCSD Birds 200-2011 dataset contains 200 different bird classes. It includes 5994 training images and 5794 testing images in total. 
 \item \textbf{StanfordDogs~\cite{KhoslaYaoJayadevaprakashFeiFei_FGVC2011}}: This dataset contains 120 different types of dogs and includes 20,580 images in total. 
\squishend
For both datasets, we only use the class labels of images in the training stage.

We choose LIBLINEAR~\cite{Fan2008JMLR_liblinear} to learn a linear
SVM classifier for classification. All the experiments are run on
a computer with Intel i7-3930K CPU and 64G main memory.

\subsection{Influences of different modules}

We investigate the effect of different modules in the proposed image representation on the CUB 200-2011 dataset in Table~\ref{table_module_acc}.

\begin{table}
 \caption{Evaluation of different modules in the proposed image representation.} \label{table_module_acc}
 \centering
 \begin{tabular}{|l|c|}
   \hline
                            & Accuracy (\%)  \\ \hline
  \textsc{conv5}+MMP+ScPM   &  71.04  \\
  \textsc{conv5}+MMP        &  68.78   \\
  \textsc{conv5}            &  58.15    \\ \hline
 \end{tabular}
\end{table}

First, we consider the effect of MMP in the proposed image representation. We compare the part proposals generated using the outputs of \textsc{conv5} and \textsc{conv5}+MMP. All part proposals are encoded into one FV in each image (not using ScPM). It can be seen that multi-scale part proposals (\textsc{conv5}+MMP) can greatly improve the recognition accuracy over single-scale part proposals (\textsc{conv5}) by $10.63\%$. This is because MMP can provide very dense coverage of object parts on different scales.

Second, we evaluate the influence of ScMP in the proposed image representation. Using the multi-scale part proposals generated by MMP, ScMP has a better accuracy (2.26\% higher) than that of the method encoding all part proposals altogether. This shows that it is beneficial to encode objects at different scales separately.

Up to now, MMP+ScMP has shown better accuracy than the state-of-the-art annotation-free fine-grained categorization method~\cite{Tianjun2014arxiv_fine2cnn} by $1.34\%$. Next, we are going to further improve the accuracy with part selection on this representation.

\subsection{Part selection}

We show the classification accuracy using part selection on the proposed image representation (MMP+ScMP) for CUB 200-2011 in Table~\ref{table_acc_CUB200_2011}.

\begin{table}
 \caption{Classification accuracy on Caltech-UCSD Birds 200-2011.} \label{table_acc_CUB200_2011}
 \centering
 \begin{tabular}{|c|c|r|}
  \hline
  \multicolumn{3}{|c|}{Without annotations in neither training nor testing} \\ \hline
  Methods  & Selection fraction      & Acc. (\%)   \\ \hline
  \multirow{5}{*}{Proposed}
    & 100\% (All) & 71.04  \\
    & 75.0\% (3/4)   & 71.67   \\
    & 50.0\% (1/2) &   73.34 \\
    & 25.0\% (1/4)   &  \textbf{75.02}  \\
    & 12.5\% (1/8)   &    73.82 \\ \hline
  \multicolumn{2}{|l|}{Two-level attention~\cite{Tianjun2014arxiv_fine2cnn}}  & 69.70  \\ \hline \hline
  \multicolumn{3}{|c|}{Use annotations in training, not in testing} \\ \hline
  \multicolumn{2}{|l|}{DPD+DeCAF~\cite{Donahue2014ICML_decaf}}  &  44.94  \\
  \multicolumn{2}{|l|}{Part based R-CNN (without parts)~\cite{ning2014ECCV_cnnFine-grained}} & 52.38 \\
  \multicolumn{2}{|l|}{Part based R-CNN-ft (without parts)~\cite{ning2014ECCV_cnnFine-grained}} & 62.75 \\
  \multicolumn{2}{|l|}{CL-45C (without parts)~\cite{liu2015CVPR_crosslayer}} & 73.50 \\
  \multicolumn{2}{|l|}{Part based R-CNN-ft (with parts)~\cite{ning2014ECCV_cnnFine-grained}} & 73.89 \\
  \multicolumn{2}{|l|}{Pose Normalized CNN~\cite{Branson2014arXiv_posecnn}} & 75.70  \\ \hline
 \end{tabular}
\end{table}

Part selection can greatly improve the accuracy. The accuracy is shown when different fractions of part clusters are selected in the image representation. When a quarter of most important part clusters (fraction 25\%) are used, a peak is reached, and it is better than that without part selection (fraction 100\%) by $3.98\%$. Even when fewer part proposals are selected (fraction 12.5\%), its accuracy is still better than that without part selection by $2.78\%$. This shows that part selection can efficiently resist the noise introduced by those part proposals from background clutters.

Our best accuracy (75.02\%) outperforms the state-of-the-art annotation-free method~\cite{Tianjun2014arxiv_fine2cnn} by $5.32\%$, and is also better than most existing annotation-dependent works. \cite{Tianjun2014arxiv_fine2cnn} claims better results can be acquired if more powerful CNN is used. For fair comparison, we cite their results using the standard CNN structure (containing 5 convolutional layers). We only show the accuracy of annotation-dependent methods using object / part annotations in the training stage, which uses the least annotations and is most close to our annotation-free setup. Most of these methods try to learn expensive part detectors to get accurate matching for recognition. However, our method shows that they are not always necessary, especially in annotation-free fine-grained categorization.

Part selection is more important in fine-grained categorization than (feature) selection in general image categorization. With part selection, the accuracy is $3.98\%$ higher than the original image representation. In~\cite{me2014cvpr_fs}, feature selection is used to compress FV for general image recognition like object recognition. Much smaller (around 1\%) improvement after selection (worse in most time)  is achieved over original FV, which is much less than the improvement in Table~\ref{table_acc_CUB200_2011}. This fact clearly shows the distinction between these two applications. In annotation-free fine-grained tasks, selecting proper object parts are critical, while in general image recognition the global image representation without selection is already good.

We show the categorization accuracy for Stanford Dogs in Table~\ref{table_StanfordDogs_acc}. The proposed method (either with or without part selection) shows much better accuracy than existing annotation-dependent works. Part selection also shows to play an important role in the proposed image representation, which leads to a $2.69\%$ improvement over the original representation. Stanford Dogs is a subset in ImageNet. It is also evaluated in~\cite{Tianjun2014arxiv_fine2cnn}, which gets worse result than ours.

\begin{table}
 \caption{Classification accuracy on StanfordDogs.} \label{table_StanfordDogs_acc}
 \centering
 \begin{tabular}{|c|c|r|}
  \hline
  \multicolumn{3}{|c|}{Without annotations in neither training nor testing} \\ \hline
  Methods  & Selection fraction      & Acc. (\%)   \\ \hline
  \multirow{5}{*}{ Proposed}
   & 100\% (All)    &  77.23  \\
   & 75.0\% (3/4)   &  78.28  \\
   & 50.0\% (1/2)   &   79.36 \\
   & 25.0\% (1/4)   &  \textbf{79.92}  \\
   & 12.5\% (1/8)   &  78.18   \\
\hline
\multicolumn{2}{|l|}{Two-level attention~\cite{Tianjun2014arxiv_fine2cnn}}  & 71.90  \\
\hline \hline
\multicolumn{3}{|c|}{Use annotations in both training and testing} \\
\hline
\multicolumn{2}{|l|}{Edge templates~\cite{Shulin2012NIPS_TemplateFineGrained}}   &   38.00  \\
\multicolumn{2}{|l|}{Unsupervised alignments~\cite{Gavves2013ICCV_finealign}}   &   50.10  \\
\multicolumn{2}{|l|}{MTL~\cite{Jian2014ECCV_MTL}}   &   39.30  \\
\hline
\end{tabular}
\end{table}

Overall, these results show that: 1) part selection is important in annotation-free fine-grained categorization; 2) it is not always necessary to learn expensive object / part detectors in fine-grained categorization.

\subsection{Key part visualization}

\begin{figure*}[p]
 \centering
 \subfloat[Red-bellied Woodpecker vs. Red-headed Woodpecker]
 {
   \includegraphics[width=0.99\textwidth]{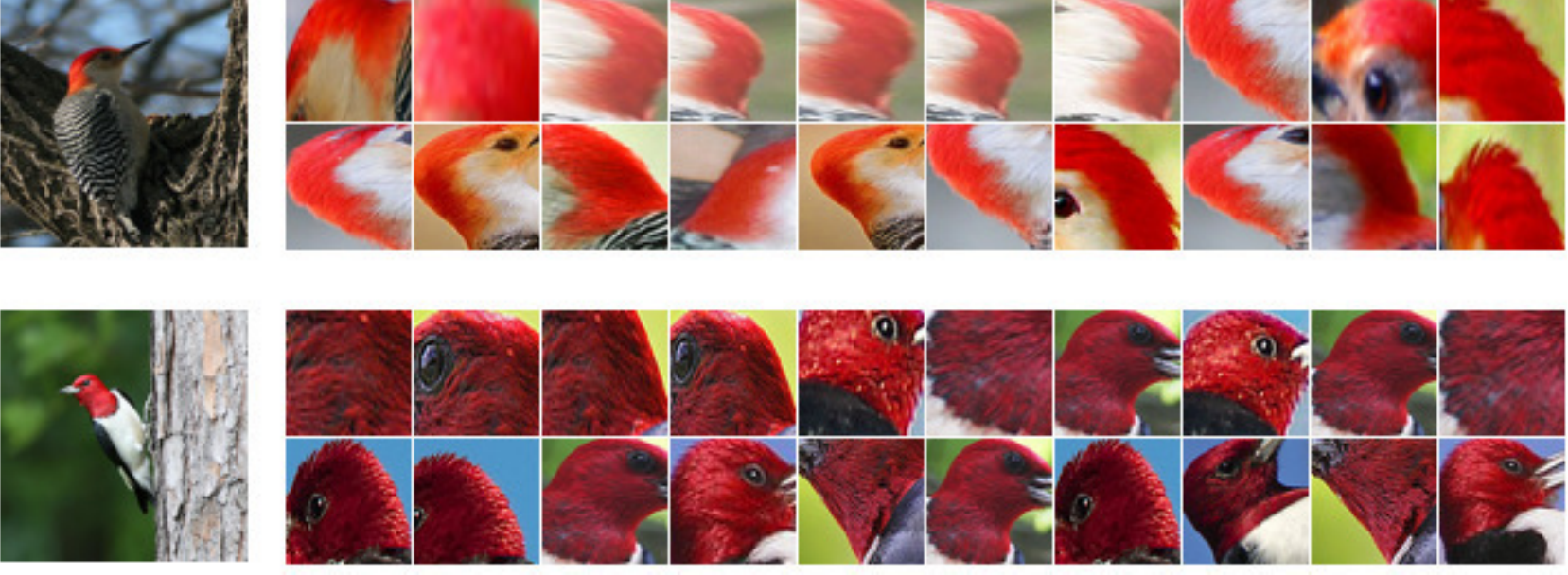}
 }
 \\
 \subfloat[Red-winged Blackbird vs. Yellow-headed Blackbird]
 {
   \includegraphics[width=0.99\textwidth]{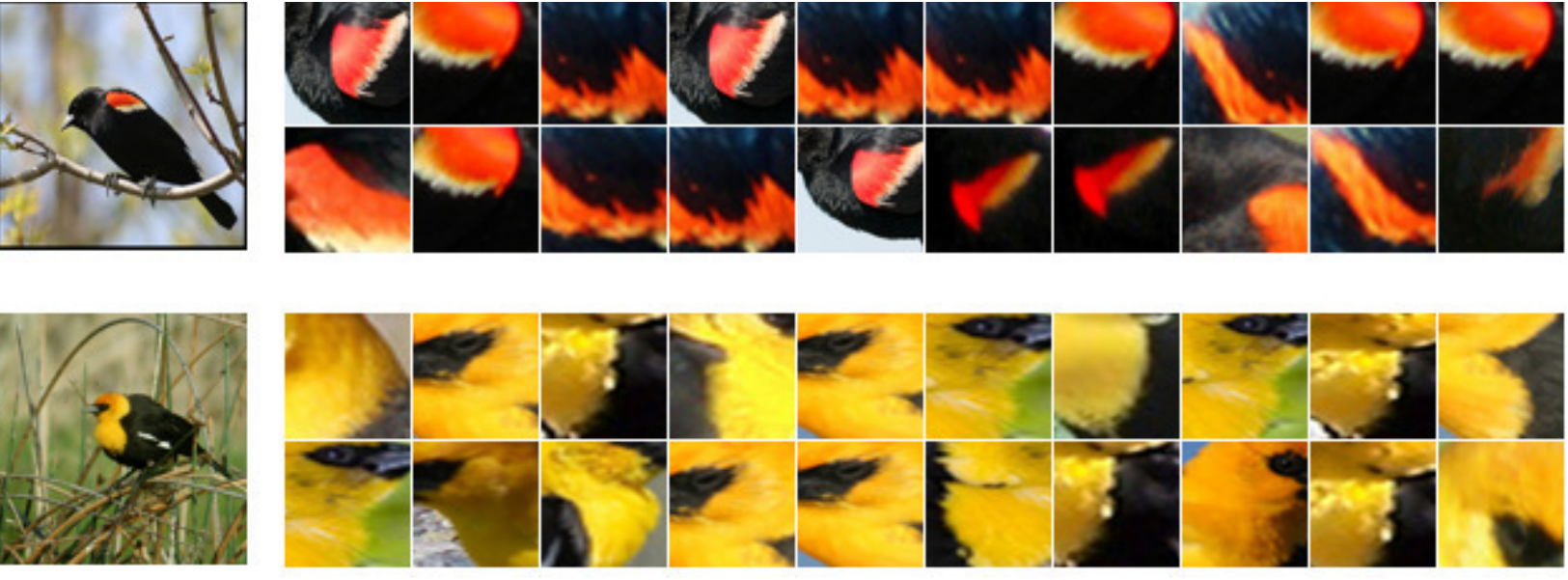}
 }
 \\
 \subfloat[Blue Jay vs. Green Jay]
 {
   \includegraphics[width=0.99\textwidth]{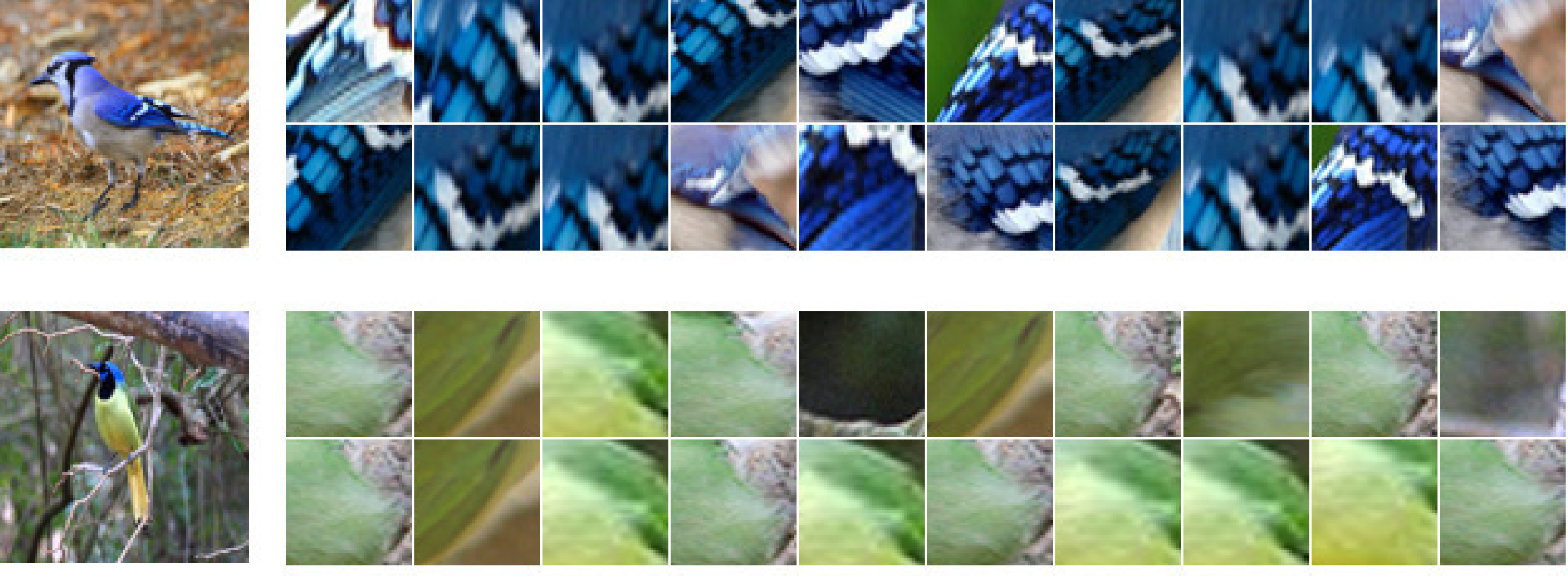}
 }
 \caption{Key (most discriminative) parts visualization for pairwise classes. Key parts are detected from testing images using the classifier learned from training images. Top 20 key parts are shown for each class. The important parts found by the proposed method coincide well with the rules human experts use to distinguish these birds. This figure is best viewed in color.} \label{fig_show_key_parts}
\end{figure*}

We detect and visualize the key parts for pairwise classes using the proposed image / part representation in Fig.~\ref{fig_show_key_parts}. In each pair, we show one sample image and 20 detected  key parts with highest (smallest) scores for the positive (negative) class. The bird names are given in the captions, which also shows how humans distinguish the two birds.

The detected parts can capture the key parts in these species, which coincides well with the human-defined rules. We also find that the proposed method can capture some tiny distinction that we cannot easily discriminate by eyes. For example, in the first pair, the key parts in the red-bellied woodpecker and red-headed woodpecker are both red, and the locations are very close. From the detected parts, we can find that the red color of red-headed woodpecker is darker and the feather of red-bellied woodpecker is finer.

From the detected parts, we can see the necessity to select many useful parts in the proposed image representation. One (best) part may cause possible loss of useful information in characterizing an object. Multiple (good) parts can compensate each other from different aspects like location, view, and scale, etc. This also explains why the proposed representation works better than~\cite{Tianjun2014arxiv_fine2cnn}, which only use detected best part for categorization.

\section{Conclusions}

In this paper, we propose to categorize fine-grained images without using any object / part annotation neither in the training nor in the testing stage. Our basic idea is to select multiple useful parts from multi-scale part proposals and use them to compute a global image representation for categorization. This is specially designed for fine-grained categorization in this annotation-free scenario, because parts have shown to play an important role in existing annotation-dependent works and accurate part detectors are hardly acquired. Particularly, we propose an efficient multi-max pooling strategy to generate multi-scale part proposals by using the internal outputs of CNN on object proposals in each image. Then, we select useful parts from those part clusters which are important for categorization. Finally, we encode the selected parts on different scales separately in a global image representation. With the proposed image / part representation technique, we use it to detect the key parts in objects of different classes, whose visualization results are intuitive and coincide well with rules used by human experts.

In the experiments, on two challenging datasets (the CUB 200-2011 and the StanfordDogs datasets), our proposed annotation-free method achieves classification accuracy of 75.02\% and 79.92\% respectively, which is better than the results of state-of-the-art annotation-free work and most existing annotation-dependent methods. Future works include utilizing the objection information mined from the global image representation to help localize objects and further improve classification.

{\small
\bibliographystyle{ieee}
\bibliography{egbib}

\begin{thebibliography}{10}\itemsep=-1pt

\bibitem{Berg2014CVPR_Birdsnap}
T.~Berg, J.~Liu, S.~W. Lee, M.~L. Alexander, D.~W. Jacobs, and P.~N. Belhumeur.
\newblock Birdsnap: Large-scale fine-grained visual categorization of birds.
\newblock In {\em CVPR}, pages 2019 -- 2026, 2014.

\bibitem{Bourdev2010ECCV_Poselet}
L.~Bourdev, S.~Maji, T.~Brox, and J.~Malik.
\newblock Detecting people using mutually consistent poselet activations.
\newblock In {\em ECCV}, volume LNCS 6316, pages 168--181, 2010.

\bibitem{Branson2014arXiv_posecnn}
S.~Branson, G.~V. Horn, S.~Belongie, and P.~Perona.
\newblock Bird species categorization using pose normalized deep convolutional
  nets.
\newblock In {\em BMVC}, 2014.

\bibitem{Chatfield2014BMVC_CNN}
K.~Chatfield, K.~Simonyan, A.~Vedaldi, and A.~Zisserman.
\newblock Return of the devil in the details: Delving deep into convolutional
  nets.
\newblock In {\em BMVC}, 2014.

\bibitem{Jia2013CVPR_FineCrowdsourcing}
J.~Deng, J.~Krause, and L.~Fei-Fei.
\newblock Fine-grained crowdsourcing for fine-grained recognition.
\newblock In {\em CVPR}, pages 580 -- 587, 2013.

\bibitem{Donahue2014ICML_decaf}
J.~Donahue, Y.~Jia, O.~Vinyals, J.~Hoffman, N.~Zhang, E.~Tzeng, and T.~Darrell.
\newblock {DeCAF}: A deep convolutional activation feature for generic visual
  recognition.
\newblock In {\em ICML}, pages 647--655, 2014.

\bibitem{Fan2008JMLR_liblinear}
R.-E. Fan, K.-W. Chang, C.-J. Hsieh, X.-R. Wang, and C.-J. Lin.
\newblock {LIBLINEAR}: A library for large linear classification.
\newblock {\em JMLR}, 9:1871--1874, Aug 2008.

\bibitem{Farrell2011ICCV_Birdlets}
R.~Farrell, O.~Oza, N.~Zhang, V.~I. Morariu, T.~Darrell, and L.~S. Davis.
\newblock Birdlets: Subordinate categorization using volumetric primitives and
  pose-normalized appearance.
\newblock In {\em ICCV}, pages 161--168, 2011.

\bibitem{Felzenszwalb2010PAMI_DPM}
P.~F. Felzenszwalb, R.~B. Girshick, D.~McAllester, and D.~Ramanan.
\newblock Object detection with discriminatively trained part based models.
\newblock {\em TPAMI}, 32:1627--1645, 2010.

\bibitem{Gavves2013ICCV_finealign}
E.~Gavves, B.~Fernando, C.~Snoek, A.~Smeulders, and T.~Tuytelaars.
\newblock Fine-grained categorization by alignments.
\newblock In {\em ICCV}, pages 1713--1720, 2013.

\bibitem{Girshick2014CVPR_RCNN}
R.~Girshick, J.~Donahue, T.~Darrell, and J.~Malik.
\newblock Rich feature hierarchies for accurate object detection and semantic
  segmentation.
\newblock In {\em CVPR}, pages 580--587, 2014.

\bibitem{Yunchao2014ECCV_cnnvlad}
Y.~Gong, L.~Wang, R.~Guo, and S.~Lazebnik.
\newblock Multi-scale orderless pooling of deep convolutional activations
  features.
\newblock In {\em ECCV}, volume LNCS 8695, pages 392--407, 2014.

\bibitem{Christoph2014CVPR_Part}
C.~Goring, E.~Rodner, A.~Freytag, and J.~Denzler.
\newblock Nonparametric part transfer for fine-grained recognition.
\newblock In {\em CVPR}, pages 2489--2496, 2014.

\bibitem{Jegou2012PAMI_VLAD}
H.~Jegou, F.~Perronnin, M.~Douze, J.~Sanchez, P.~Perez, and C.~Schmid.
\newblock Aggregating local image descriptors into compact codes.
\newblock {\em TPAMI}, 34(9):1704--1716, 2012.

\bibitem{Yangqing2012CVPR_ReceptiveField}
Y.~Jia, C.~Huang, and T.~Darrell.
\newblock Beyond spatial pyramids: Receptive field learning for pooled image
  features.
\newblock In {\em CVPR}, pages 3370--3377, 2012.

\bibitem{KhoslaYaoJayadevaprakashFeiFei_FGVC2011}
A.~Khosla, N.~Jayadevaprakash, B.~Yao, and L.~Fei-Fei.
\newblock Novel dataset for fine-grained image categorization.
\newblock In {\em First Workshop on Fine-Grained Visual Categorization, CVPR},
  2011.

\bibitem{Lazebnik2006CVPR_SPM}
S.~Lazebnik, C.~Schmid, and J.~Ponce.
\newblock Beyond bags of features: Spatial pyramid matching for recognizing
  natural scene categories.
\newblock In {\em CVPR}, 2006.

\bibitem{liu2015CVPR_crosslayer}
L.~Liu, C.~Shen, and A.~van~den Hengel.
\newblock The treasure beneath convolutional layers: Cross-convolutional-layer
  pooling for image classification.
\newblock In {\em CVPR}, 2015.

\bibitem{Nilsback2008ICVGIP_flowers}
M.-E. Nilsback and A.~Zisserman.
\newblock Automated flower classification over a large number of classes.
\newblock In {\em ICVGIP}, pages 722--729, 2008.

\bibitem{Wanli2014arxiv_deepidnet}
W.~Ouyang, P.~Luo, X.~Zeng, S.~Qiu, Y.~Tian, H.~Li, S.~Yang, Z.~Wang, Y.~Xiong,
  C.~Qian, Z.~Zhu, R.~Wang, C.-C. Loy, X.~Wang, and X.~Tang.
\newblock {DeepID-Net}: multi-stage and deformable deep convolutional neural
  networks for object detection.
\newblock In {\em CVPR}, 2015.

\bibitem{Jian2014ECCV_MTL}
J.~Pu, Y.-G. Jiang, J.~Wang, and X.~Xue.
\newblock Which looks like which: Exploring inter-class relationships in
  fine-grained visual categorization.
\newblock In {\em ECCV}, volume LNCS 8691, pages 425--440, 2014.

\bibitem{Sanchez2013IJCV_FV}
J.~S\'anchez, F.~Perronnin, T.~Mensink, and J.~Verbeek.
\newblock Image classification with the {Fisher} vector: Theory and practice.
\newblock {\em IJCV}, 105(3):222--245, 2013.

\bibitem{Rejeb2013CVPR_VantageFine}
A.~R. Sfar, N.~Boujemaa, and D.~Geman.
\newblock Vantage feature frames for fine-grained categorization.
\newblock In {\em CVPR}, pages 835--842, 2013.

\bibitem{Uijlings2013IJCV_SS}
J.~Uijlings, K.~van~de Sande, T.~Gevers, and A.~Smeulders.
\newblock Selective search for object recognition.
\newblock {\em IJCV}, 104:154--171, 2013.

\bibitem{Vedaldi2014CVPR_fineplanes}
A.~Vedaldi, S.~Mahendran, S.~Tsogkas, S.~Maji, B.~Girshick, J.~Kannala,
  E.~Rahtu, I.~Kokkinos, M.~B. Blaschko, D.~Weiss, B.~Taskar, K.~Simonyan,
  N.~Saphra, and S.~Mohamed.
\newblock Understanding objects in detail with fine-grained attributes.
\newblock In {\em CVPR}, pages 3622--3629, 2014.

\bibitem{WahCUB_200_2011}
C.~Wah, S.~Branson, P.~Welinder, P.~Perona, and S.~Belongie.
\newblock {The Caltech-UCSD Birds-200-2011 Dataset}.
\newblock Technical Report CNS-TR-2011-001, California Institute of Technology,
  2011.

\bibitem{Catherine2014CVPR_Interactivefine}
C.~Wah, G.~V. Horn, S.~Branson, S.~Maji, P.~Perona, and S.~Belongie.
\newblock Similarity comparisons for interactive fine-grained categorization.
\newblock In {\em CVPR}, pages 859--866, 2014.

\bibitem{Tianjun2014arxiv_fine2cnn}
T.~Xiao, Y.~Xu, K.~Yang, J.~Zhang, Y.~Peng, and Z.~Zhang.
\newblock The application of two-level attention models in deep convolutional
  neural network for fine-grained image classification.
\newblock In {\em CVPR}, 2015.

\bibitem{Shulin2012NIPS_TemplateFineGrained}
S.~Yang, L.~Bo, J.~Wang, and L.~Shapiro.
\newblock Unsupervised template learning for fine-grained object recognition.
\newblock In {\em NIPS 26}, pages 3131--3139, 2012.

\bibitem{Donggeun2014arxiv_fvcnn}
D.~Yoo, S.~Park, J.-Y. Lee, and I.~S. Kweon.
\newblock Fisher kernel for deep neural activations.
\newblock arXiv:1412.1628v2, 2014.

\bibitem{Zeiler2014ECCV_visualizeCNN}
M.~D. Zeiler and R.~Fergus.
\newblock Visualizing and understanding convolutional networks.
\newblock In {\em ECCV}, volume LNCS 8689, pages 818--833, 2014.

\bibitem{ning2014ECCV_cnnFine-grained}
N.~Zhang, J.~Donahue, R.~Girshick, and T.~Darrell.
\newblock Part-based {R-CNN}s for fine-grained category detection.
\newblock In {\em ECCV}, volume LNCS 8689, pages 834--849, 2014.

\bibitem{Ning2012CVPR_posepooling}
N.~Zhang, R.~Farrell, and T.~Darrell.
\newblock Pose pooling kernels for sub-category recognition.
\newblock In {\em CVPR}, pages 3665--3672, 2012.

\bibitem{Ning2013CVPR_DPD}
N.~Zhang, R.~Farrell, F.~Iandola, and T.~Darrell.
\newblock Deformable part descriptors for fine-grained recognition and
  attribute prediction.
\newblock In {\em CVPR}, pages 729--736, 2013.

\bibitem{Ning2014CVPR_PANDA}
N.~Zhang, M.~Paluri, M.~Ranzato, T.~Darrell, and L.~Bourdev.
\newblock {PANDA}: Pose aligned networks for deep attribute modeling.
\newblock In {\em CVPR}, pages 1637--1644, 2014.

\bibitem{me2014cvpr_fs}
Y.~Zhang, J.~Wu, and J.~Cai.
\newblock Compact representation for image classification: To choose or to
  compress?
\newblock In {\em CVPR}, pages 907--914, 2014.

\end{thebibliography}
}
\end{document}